\documentclass{article}

\usepackage[preprint,nonatbib]{neurips_2021}

\usepackage[utf8]{inputenc} 
\usepackage[T1]{fontenc}    
\usepackage{hyperref}       
\usepackage{url}            
\usepackage{booktabs}       
\usepackage{amsfonts}       
\usepackage{nicefrac}       
\usepackage{microtype}      
\usepackage{xcolor}         

\usepackage{wrapfig}
\usepackage{graphicx}
\usepackage{placeins}
\usepackage{subcaption}
\usepackage{amsmath}
\usepackage{amsthm}

\usepackage[subtle]{savetrees}

\newtheorem{theorem}{Theorem}
\newtheorem{lemma}{Lemma}

\title{On Local Aggregation in Heterophilic Graphs}

\author{%
  Hesham Mostafa\\
  Intel Labs\\
  San Diego, United States\\
  \texttt{hesham.mostafa@intel.com} \\
  \And
  Marcel Nassar \\
  Intel Labs \\
  San Diego, United States \\
  \texttt{marcel.nassar@intel.com} \\
  \And
  Somdeb Majumdar \\
  Intel Labs \\
  San Diego, United States \\
  \texttt{somdeb.majumdar@intel.com} \\
}

\begin{document}

\maketitle

\begin{abstract}
Many recent works have studied the performance of Graph Neural Networks (GNNs) in the context of graph \textit{homophily} - a label-dependent measure of connectivity. Traditional GNNs generate node embeddings by aggregating information from a node's neighbors in the graph. Recent results in node classification tasks show that this local aggregation approach performs poorly in graphs with low homophily (heterophilic graphs). Several mechanisms have been proposed to improve the accuracy of GNNs on such graphs by increasing the aggregation range of a GNN layer, either through multi-hop aggregation, or through long-range aggregation from distant nodes.
In this paper, we show that properly tuned classical GNNs and multi-layer perceptrons match or exceed the accuracy of recent long-range aggregation methods on heterophilic graphs. Thus, our results highlight the need for alternative datasets to benchmark long-range GNN aggregation mechanisms. We also show that homophily is a poor measure of the information in a node's local neighborhood and propose the \textit{Neighborhood Information Content} (NIC) metric, which is a novel  information-theoretic graph metric.
We argue that NIC is more relevant for local aggregation methods as used by GNNs. We show that, empirically, it correlates better with GNN accuracy in node classification tasks than homophily.
\end{abstract}

\section{Introduction}
\label{sec:intro}
Graph neural networks (GNNs) are networks  which operate on graph-structured data ~\cite{Kipf_Welling16}. They achieve excellent performance on a large variety of problems such as node classification~\cite{Shi_etal20,Li_etal20}, graph classification~\cite{Corso_etal20}, graph matching~\cite{Li_etal19}, and link prediction~\cite{Zhang_Chen18}. A layer in a vanilla GNN generates an output feature vector for each node by aggregating input features from its neighboring nodes in the graph. This local aggregation mechanisms has multiple advantages: it is a natural method of making the output feature vectors depend on the graph structure. It also makes GNNs more scalable as a node does not need to consider the entire graph, but only its local neighborhood to generate its output feature vector.

 In node classification problems, a recent common graph metric is label-based graph homophily~\cite{Newman_03} which is the fraction of edges that connect two nodes with the same label. Previous work shows that traditional GNNs with local aggregation perform poorly on graphs with low homophily~\cite{Zhu_etal20,Pei_etal20}, also known as heterophilic graphs. An underlying assumption in traditional, locally aggregating GNNs, is that the most relevant information for classifying or embedding a node is contained in its graph neighborhood. At first sight, heterophilic graphs seem to violate this assumption as a node's immediate neighborhood predominantly contains nodes from other classes, which might explain why GNNs with local aggregation perform poorly on heterophilic graphs.

 This perceived limitation of GNNs motivated the search for methods that would allow a node to directly aggregate information from beyond its graph neighborhood. One solution involves allowing each node to directly aggregate information from nodes that are multiple hops away, and not just from the node's immediate neighborhood~\cite{abu_etal19,Zhu_etal20}. An alternative solution, which we term the long-range aggregation solution, allows a node to potentially aggregate information from any other node in the graph. Long range aggregation methods have to contend with memory and compute scalability issues as each node would need to consider a larger set of input nodes, and not just its local neighborhood. They also need extra heuristics to find for each node the most relevant distant nodes. Several long-range aggregation mechanisms have been proposed that address these two concerns~\cite{Pei_etal20,Mostafa_Nassar20,Liu_etal20} and they demonstrate superior accuracy on various heterophilic datasets compared to simple local aggregation.

 In this paper, our contribution is two-fold: 
 \begin{enumerate}
     \item We show that classical GNNs and Multi-layer Perceptrons (MLPs) with properly tuned hyper-parameters match or exceed the performance of recently proposed multi-hop and long-range aggregation methods on  synthetic and real-world heterophilic datasets. This indicates that current heterophilic datasets do not offer a conclusive means for showing the advantages of multi-hop or long-range aggregation over the simple one-hop aggregation in vanilla GNNs. 
    \item We introduce a new graph metric, the \textit{Neighborhood Information Content} (NIC) metric, that estimates the amount of information contained in a node's neighborhood that is relevant for predicting the node's label. We formulate this metric as a lower bound on the mutual information between the label of a node and the labels of its neighbors.  We calculate this metric for real-world and synthetic datasets and show that it is a better predictor of the accuracy of vanilla GNNs on node classification tasks than homophily.
\end{enumerate}

\section{Related Work}
\label{sec:related}
Early local aggregation methods used in GNNs were motivated by spectral convolutions~\cite{Defferrard_etal16}, in particular by first order spectral convolutions. These can be approximated by localized 1-hop convolutions in the spatial (non-spectral) domain~\cite{Kipf_Welling16}. A large variety of localized aggregation mechanisms have since been developed that broke away from spectral methods in favor of more expressive spatial aggregation methods~\cite{Velivckovic_etal17,Hamilton_etal17,Xu_etal18,Simonovsky_Komodakis17}. These methods, however, still generate new node features by aggregating information from a node's immediate neighborhood. One-hop aggregation methods can be stacked together, often separated by non-linearities, to increase the aggregation range of each node. This, however, is not always successful in capturing information from far-away nodes as information can be aggregated from too many nodes, drowning out relevant contributions. This is the over-smoothing phenomenon~\cite{Li_etal18b,Chen_etal19} which, for excessively large aggregation ranges, can produce output features that are very similar across the different nodes since the aggregation ranges of the different nodes overlap strongly.

Instead of stacking multiple one-hop aggregation layers, multi-hop aggregation can be implemented by directly considering the $k^{th}$ power of the adjacency matrix while doing a single propagation step which would effectively aggregate information directly from nodes up to $k$-hops away~\cite{abu_etal19}. Random walks with teleport can also be used to define a node's aggregation range~\cite{Klicpera_etal18}. These methods are still not true long-range aggregation methods as they can not aggregate information from nodes that are arbitrarily far away in the graph.

Another class of aggregation methods, which we term long-range aggregation methods, are not strictly tied to the graph structure and thereby allow a node to potentially aggregate information from any other node in the graph. Geom-GCN~\cite{Pei_etal20} uses standard node embedding methods and aggregates information from nearby nodes in the embedding space. The embeddings are pre-computed and are not learned when training the GNN. Permutohedral-GCNs~\cite{Mostafa_Nassar20} use learnable embeddings and a global attention mechanism to aggregate information from all nodes weighted by their distance from the target node in the embedding space. Non-local graph neural networks project trainable node embeddings onto a $1$D line to generate an ordering for the graph nodes, then use $1$D convolution to allow each node to aggregate features from its neighbors on the $1$D line~\cite{Liu_etal20}.

The main motivation of long-range aggregation methods is to allow GNNs to handle heterophilic graphs~\cite{Newman_03}. Recently, the notion of homophily and heterophily has become strongly associated with the class label of nodes~\cite{Pei_etal20}; long-range aggregation methods argue that for node classification in graphs with small label-based homophily, the local aggregation of vanilla GNNs is insufficient to learn the class label of nodes as a node will be predominantly aggregating information from nodes with different labels~\cite{Zhu_etal20,Pei_etal20,Mostafa_Nassar20,Liu_etal20}. 

\section{Background}

Unless stated otherwise, We use the term GNNs to denote graph networks that use a local aggregation mechanism. We use two standard variants of GNNs: Graph Convolutional Networks (GCNs)~\cite{Kipf_Welling16} and GraphSage~\cite{Hamilton_etal17}. Given a directed graph $G(\mathcal{V},\mathcal{E})$ where $\mathcal{V}$ is a set of $N$ nodes/vertices and $\mathcal{E} \subset \mathcal{V} \times \mathcal{V}$ is the set of edges.  $\mathcal{N}(i)$ is the neighborhood of node $i$, and $|\mathcal{N}(i)|$ is the degree of node $i$. Let ${\bf h}^{k}_i \in \mathbb{R}^{F_{k}}$ be the output feature vector of node $i$ at layer $k$ where $F_{k}$ is the feature dimension at layer $k$. A  GCN layer produces the node output feature vectors at layer $k$ according to:
\begin{equation}
\label{eq:gcn}  
{\bf h}_i^{k} = ReLU\left({\bf b}^{k} + \sum\limits_{j\in \mathcal{N}(i)} \frac{1}{\sqrt{|\mathcal{N}(i)||\mathcal{N}(j)|}}{\bf W}^k{\bf h}_j^{k-1}\right),
\end{equation}
where ${\bf b}^{k} \in \mathbb{R}^{F_k} $ and ${\bf W}^k \in \mathbb{R}^{F_k\times F_{k-1}}$ are the layer's learnable parameters. GraphSage extends the node aggregation mechanism to treat the node's own features differently from the features of its neighbors. We use the GraphSage layer with mean aggregation defined by:
\begin{equation}
  {\bf h}_i^{k} = ReLU\left({\bf b}^{k} + {\bf W}^{k}\left[{\bf h}_i^{k-1}||\frac{1}{|\mathcal{N}(i)|}\sum\limits_{j\in \mathcal{N}(i)} {\bf h}_j^{k-1}\right]\right),
  \label{eq:gsage}
\end{equation}
where $||$ is the concatenation operator, and ${\bf b}^{k} \in \mathbb{R}^{F_k} $ and ${\bf W}^k \in \mathbb{R}^{F_k\times 2F_{k-1}}$ are the layer's learnable parameters. We use the full-batch GraphSage, i.e, we consider the node's entire neighborhood without any neighborhood sampling. We also use vanilla MLPs that do not take the graph structure into account, i.e, they operate on each node's feature vector independently of other nodes. We use a standard MLP layer:
\begin{equation}
{\bf h}_i^{k} = ReLU\left({\bf b}^{k} + {\bf W}^{k}{\bf h}_i^{k-1}\right),
\end{equation}
where ${\bf b}^{k} \in \mathbb{R}^{F_k} $ and ${\bf W}^k \in \mathbb{R}^{F_k\times F_{k-1}}$ are the MLP layer's learnable parameters. 

Graph homophily is a measure of the tendency of similar nodes in the graph to connect to each other. There could be different measures of node similarity that depend on the different attributes associated with each node~\cite{Newman_03}. In the context of node classification, similarity is often defined based on the node labels, and is a binary quantity that depends on whether two nodes belong to the same class or not: $s(i,j) = 1$ if nodes $i$ and $j$ have the same label and $s(i,j) = 0$ otherwise . Given this binary label-based node similarity measure, there are two often-used types of graph homophily: node-wise homophily~\cite{Pei_etal20,Liu_etal20} and edge-wise homophily~\cite{Zhu_etal20,abu_etal19} which are defined, respectively, as:
\begin{equation}
\mathcal{H}_{node} = \frac{1}{N} \sum\limits_{i=1}^N \frac{1}{|\mathcal{N}(i)|} \sum\limits_{j \in \mathcal{N}(i)} s(i,j)
\end{equation}

\begin{equation}
\mathcal{H}_{edge} = \frac{1}{|\mathcal{E}|} \sum\limits_{(i,j)\in \mathcal{E}}s(i,j) = \frac{1}{|\mathcal{E}|}\sum\limits_{i=1}^N \sum\limits_{j \in \mathcal{N}(i)} s(i,j)
\end{equation}
Assortivity is another popular metric that has been used in network science to characterize  mixing patterns in graphs \cite{Newman_03}. Assortivity is given by:
\begin{equation}
    r = \frac{Tr\{\mathbf{E}\} - \parallel \mathbf{E}\parallel^2}{1 - \parallel\mathbf{E}\parallel^2}
\end{equation}
where $Tr\{\cdot\}$ is the matrix trace and $\parallel\cdot\parallel$ is the Frobenius norm. $\mathbf{E}$ is a matrix whose element $e_{ij}$ is the fraction of edges in a network that connect a vertex of type $i$ to a vertex of type $j$.
These notions of homophily are subtly different. In this paper, we use the edge-wise definition of homophily $\mathcal{H}_{edge}$ and observe that our results hold for the other measures which either give very similar values, or values that change in a linear fashion with $\mathcal{H}_{edge}$ .

\section{The Neighborhood Information Content (NIC) Metric}
\label{sec:nic}
In node classification tasks, label-based homophily has often been used as a measure of how well GNNs are likely to perform. However, a rigorous reason behind this common notion is lacking. In fact, as we show in the Results section, homophily is often a poor predictor of the accuracy of GNNs in node classification tasks. In this section we propose an alternative metric which directly measures how well a node's label can be predicted from the labels of its neighbors.

Let $L = \{1,\ldots,C\}$ be the set of $C$ distinct labels and $S_d = \{\{l_1,\ldots,l_d\} : l_i \in L\text{ for $i=1,\ldots,d$}\}$ be the set containing all sets of labels of size $d$. $S_d$ is thus the space of all possible label configurations in the neighborhood of a node with degree $d$. Note that the configurations are unordered, thus reflecting the permutation-invariant nature of the neighborhood. We define the set of all possible neighborhood label configurations of size up to $D$ as $S^D = \bigcup_{d = 0}^D S_d$. In a graph with maximum node degree $D$, and for a vertex $v$, let $l(v)$ denote the label of $v$ and $l(\mathcal{N}(v)) = \{l(w) : w \in \mathcal{N}(v)\}$ the label configuration of its neighborhood. We are interested in modeling the joint probability distribution $P(l(\mathcal{V}),l(\mathcal{N}(\mathcal{V})) : (L \times S^D) \rightarrow [0,1]$ which is the probability distribution over node labels and the label configurations of their neighborhoods, i.e, $P(l_x,l(\mathcal{N}_x))$ is the probability of finding a node with label $l_x$ whose neighborhood label configuration is $l(\mathcal{N}_x)$. 

We want to estimate the mutual information (MI) between the label of a node and the label configuration of its neighborhood: $MI\left(l(\mathcal{V});l(\mathcal{N}(\mathcal{V}))\right)$. Estimating MI in high dimensions is challenging. Estimation using empirical distributions obtained from the graph are bound to be inaccurate as there would not be enough nodes and neighborhood configurations to properly cover the high dimensional space. Kernel methods~\cite{Joe89} require choosing appropriate kernels and are inaccurate in data-limited situations. Recently, MI estimation using neural networks demonstrated some success in high-dimensional spaces~\cite{Belghazi_etal18}. However obtaining the estimate requires training a neural network, and the final estimate is highly sensitive to the architecture and training procedure of the estimation network~\cite{Tschannen_etal19}.

Our solution is to bound the MI under some assumptions on the graph connectivity; we assume the connection probability between any two nodes in the graph depends only on their labels. We can then prove the following theorem:
\begin{theorem}
  \label{th:main}
  \textbf{Lower bound on $MI\left(l(\mathcal{V});l(\mathcal{N}(\mathcal{V}))\right)$ under label-dependent connectivity}. Given a random graph $G(\mathcal{V},\mathcal{E})$ with labelled nodes where the following two assumptions hold:
    \begin{enumerate}
      \item the probability that a node with label $r$ connects to a node with label $s$ is $c_{sr}$.
      \item the maximum node degree is $D$, such that $N = |\mathcal{V}| \gg D$.
    \end{enumerate} 
      Let $q_s$ be the probability distribution over node degrees for nodes with label $s$, $C$  the number of distinct labels, and $p_s$ the probability that a node has label $s$, the following is a lower bound on $MI\left(l(\mathcal{V});l(\mathcal{N}(\mathcal{V}))\right)$:
  \begin{flalign*}
    NIC = -\sum\limits_{s=1}^{C}p_s ln\left(\sum\limits_{r=1}^{C}p_r \left(\sum\limits_{d=0}^{D}\sqrt{q_s(d)q_r(d)}\left(\sum\limits_{k=1}^{C}\sqrt{z_{sk}z_{rk}}\right)^d\right)\right),
  \end{flalign*}
  where
  \begin{equation*}
    z_{sr} = \frac{p_rc_{sr}}{\sum\limits_{k=1}^C p_kc_{sk}}
  \end{equation*}
  We term this lower bound the Neighborhood Information Content (NIC) metric.
\begin{proof}
The proof is given in appendix~B.
\end{proof}
\end{theorem}
  
All the quantities in the lower bound can be easily estimated in an empirical way from a given graph. $p_s$ is simply the fraction of nodes with label $s$. $c_{sr}$ is the fraction of edges present from all nodes of class $r$ to all nodes of class $s$. Alternatively, $z_{sr}$ can be estimated directly as the fraction of nodes with label $r$ present in all the neighborhoods of nodes with label $s$. $q_s$ is the empirical distribution of the degrees of nodes with label $s$. Note that $q_s$ can be derived from the quantities $c_{sr}$ and $p_s$ and does not need to be separately estimated. However, we prefer to keep the $q_s$s as independent variables to highlight the dependence of NIC on the node degree distribution. Computationally, the lower bound can be quickly evaluated as it involves only $C^3D$ terms.

Compared to homophily, our proposed information theoretic bound has several advantages:
\begin{enumerate}
    \item It takes into account inter-class connectivity patterns through the terms $c_{sr}$ unlike homophily which only considers whether a node is connected to another of the same label or not. 
    \item It takes into account the distribution of node labels through the $p_s$ terms. 
    \item It depends on the degree distribution in the graph and yields higher MI estimates for more densely connected graphs. This reflects the fact that in the presence of noise, nodes with higher degrees are able to exploit their larger neighborhoods to filter out the noise and obtain a better estimate of their labels. We demonstrate this phenomenon empirically in some of our benchmarks.
\end{enumerate}
One limitation of using NIC to gauge the potential accuracy of GNNs is that NIC depends only on the distribution of labels and not on node features. The latter is more relevant to GNN accuracy as a GNN layer typically aggregates the features of its neighbors and not their labels (see ref.~\cite{Shi_etal20} for an exception). However, different node labels typically yield different distributions of node features. Thus, NIC should be an indicator of how strongly a node's label depends on the features of its neighbors in the graph. In the next section, we empirically demonstrate that this informal argument holds true in many cases and that NIC is a better general predictor of the accuracy of GNNs with local aggregation in node classification tasks than homophily.

\section{Experimental Results}
\label{sec:results}
For all models, we use two hidden layers, and use dropout between all layers. We tune hyper-parameters to obtain the best validation accuracy for each dataset, and report test results at the best validation points. We use the ADAM optimizer throughout~\cite{Kingma_Ba14}. See Appendix~A for more details on the hyper-parameter tuning scheme. Our hyper-parameter tuning scheme has a low overhead as we only consider few possible values for each hyper-parameter (for example only 3 values for the learning rate). 
\subsection{Real-world Datasets}
We start by comparing GCNs~\cite{Kipf_Welling16}, GraphSage~\cite{Hamilton_etal17}, and MLPs against existing GNN methods designed to boost accuracy on real-world heterophilic graphs. The heterophilic graphs we use are summarized in table~\ref{tab:dataset_properties}. The graphs were obtained from ref.~\cite{Pei_etal20}. The training and evaluation procedures for these graphs are different across different papers making it hard to compare the accuracy of different approaches. The differences stem from the inconsistent training/validation/testing splits used. In Geom-GCN, the authors claim to have used a 60\%/20\%/20\% training/validation/testing split, but in the accompanying code, the split is actually 48\%/32\%/20\%. Some methods use a random 60\%/20\%/20\% split~\cite{Mostafa_Nassar20,Liu_etal20}, while other methods follow the Geom-GCN split~\cite{Zhu_etal20}. To ensure fair comparison, we repeat all our experiments twice: once with the geom-gcn split provided by the authors~\cite{Pei_etal20}, and once using a random 60\%/20\%/20 split. For the $10$ geom-gcn splits provided for each dataset, we run $10$ trials for each split for a total of $100$ training trials per dataset. For the random 60\%/20\%/20 split, we sample 20 different splits and run 5 trials per split, also resulting in 100 trials per dataset. Following ~\cite{Pei_etal20}, we add self-loops to all graphs. For the Chameleon and Squirrel datasets we make the graphs undirected.

Table~\ref{tab:main_results} shows the results for the two different split strategies, where previous methods are grouped based on the split strategy they used. For three of the datasets used: Cornell, Texas, Wisconsin, the test accuracy is so noisy across the different trials that there is no significant winner. It is clear, however, that MLPs outperform GCNs and GraphSage on these datasets. Compared to state of the art methods, the mean accuracy of MLPs is either the same or well within the standard deviation of the best performing methods on all datasets except Chameleon and Squirrel. 
These results indicate that multi-hop aggregation methods( $H_2$GCN~\cite{Zhu_etal20}) or long-range aggregation methods(Geom-GCN, PH-GCN, NLGNN~\cite{Pei_etal20,Mostafa_Nassar20,Liu_etal20}) might be superfluous for these four datasets.

For the Chameleon and Squirrel datasets with the 60\%/20\%/20\% split, we see that GraphSage outperforms non-local graph neural networks (NLGNN)~\cite{Liu_etal20}, though the advantage is hardly significant. For the Geom-GCN split, GraphSage outperforms other methods by a large margin, though we suspect that this large accuracy gain might be partly due to prior methods using the directed version of these graphs instead of the undirected version we are using. While multi-hop and long-range aggregation methods are more general than MLPs or GraphSage, table~\ref{tab:main_results} shows that they do not have an advantage on commonly used heterophilic benchmarks.

Table~\ref{tab:dataset_properties} shows the NIC metric for the different datasets. NIC is higher for the Chameleon dataset compared to the Cornell, Texas, Wisconsin, and Actor datasets. This is in line with the superior relative accuracy of local aggregation methods vs. MLPs on the Chameleon dataset. The Squirrel dataset has an anomalous NIC measurement which puts it in the same range as datasets where local aggregation has no benefits, even though local aggregation improves accuracy in the Squirrel dataset (GCNs perform better than MLPs).

\begin{table}[h]
\caption{Properties of common heterophilic graph datasets}
\label{tab:dataset_properties}
\begin{center}
\begin{tabular}{l|cccccc}
\hline
  & Chameleon & Squirrel  & Cornell & Texas & Wisconsin & Actor \\
    \hline
    \hline
    $\mathcal{H}_{edge}$ & 0.24  & 0.22 & 0.3 & 0.11 &  0.20 & 0.22 \\
    $NIC(nats)$ & 0.60  & 0.25 & 0.25 & 0.36 &  0.29 & 0.13 \\    
    Number of nodes & 2277 & 5201 & 183 & 183 & 251 & 7600\\
    Number of edges & 36101 & 217073  & 295 & 309 & 499 & 33544\\
    Node feature dimensions & 2325 & 2089 & 1703 & 1703 & 1703 & 931\\
    Number of classes & 5 & 5  & 5 & 5 & 5 & 5\\
\end{tabular}
\end{center}
\end{table}

\begin{table}[h]
\caption{Mean percentage accuracy and standard deviation on common heterophilic datasets}
\label{tab:main_results}
\begin{center}
  \small
  \addtolength{\tabcolsep}{-3pt}
\begin{tabular}{l|cccccc}
\hline
Dataset (60\%/20\%/20\% split)  & Chameleon & Squirrel & Cornell & Texas & Wisconsin & Actor \\
    \hline
    \hline
    MLP & $48.8 \pm 2.0$ & $32.3 \pm 1.1$ & ${\bf 84.9 \pm 6.1}$ & ${\bf 85.4 \pm 5.8}$ & $84.8 \pm 4.7$ & $37.0 \pm 1.1$\\
    GCN & $67.8 \pm 2.2$ & $54.4 \pm 1.4$ & $53.8 \pm 3.8$ & $54.1 \pm 2.3$ & $46.0 \pm 3.5$ & $29.0 \pm 1.0$\\
    GraphSage & ${\bf 71.0 \pm 1.8}$ & ${\bf 60.3 \pm 0.01}$ & $72.6 \pm 7.2$ & $78.7 \pm 6.6$ & $78.7 \pm 5.2$  & $34.1 \pm 1.0$   \\
    
    PH-GCN~\cite{Mostafa_Nassar20} & - & - &  $74.3 \pm 6.5$  & $63.2\pm 5.6$ & $68.2\pm 7.3$ & $34.3 \pm 1.3$ \\
    NLGNN~\cite{Liu_etal20} & $70.1 \pm 2.9$ & $59.0 \pm 1.2$ & ${\bf 84.9 \pm 5.7}$ & ${\bf 85.4 \pm 3.8}$ & ${\bf 87.3 \pm 4.3}$  & ${\bf 37.9 \pm 1.3}$   \\

    \hline
    Dataset (Geom-GCN split)  & Chameleon & Squirrel & Cornell & Texas & Wisconsin & Actor \\
    \hline
    \hline
    MLP & $48.3 \pm 2.1$ & $32.0 \pm 1.5$ & ${\bf 83.9 \pm 6.9}$ & $83.1 \pm 5.0$ & $85.3 \pm 4.6$ & ${\bf 36.1 \pm 1.1}$\\
    GCN & $66.9 \pm 1.7$ & $53.4 \pm 1.3$ & $55.2 \pm 7.6$ & $59.4 \pm 4.4$ & $46.5 \pm 8.3$ & $28.5 \pm 1.1$\\
    GraphSage & ${\bf 70.5 \pm 1.5}$ & ${\bf 57.1 \pm 1.6}$ & $70.5 \pm 6.5$ & $76.0 \pm 6.8$ & $76.7 \pm 5.6$ & $33.1 \pm 1.2$\\    
    Geom-GCN-I~\cite{Pei_etal20} & $60.31$ & $33.32$ & $56.76$ & $57.58$ & $58.24$ & $29.09$\\
    Geom-GCN-P~\cite{Pei_etal20} & $60.90$ & $38.14$ &  $60.81$ & $67.57$ & $64.12$ & $31.63$\\
    Geom-GCN-S~\cite{Pei_etal20} & $59.96$ & $36.24$ & $55.68$ & $59.73$ & $56.67$ & $30.30$\\
    $H_2$GCN-1~\cite{Zhu_etal20} & $57.11 \pm 1.58 $ & $36.42 \pm 1.89$ & $82.16 \pm 4.80$ & ${\bf 84.86 \pm 6.77}$ & ${\bf 86.67 \pm 4.69}$ & $35.86 \pm 1.03$ \\
    $H_2$GCN-2~\cite{Zhu_etal20} & $59.39 \pm 1.98$ & $37.90 \pm 2.02$ & $82.16 \pm 6.00$ & $82.16 \pm 5.28$ & $85.88 \pm 4.22$ & $35.62 \pm 1.30$ \\    
\end{tabular}
\end{center}
\end{table}

\subsection{Synthetic Datasets}
\label{sec:synth-data}
We use synthetic graphs generated using the preferential attachment method from ref.~\cite{Barabasi_Albert99}: starting from a small core graph, new nodes are sequentially added. Each new node randomly connects to a subset of the current nodes in the graph. The probability that the new node will connect to an existing node is proportional to the degree of the existing node. This rich-gets-richer connection scheme gives rise to a graph with a power law degree distribution. Graph homophily can be incorporated in the generation process by having the connection probabilities depend on the node labels in order to produce a graph with the desired homophily level~\cite{Karimi_etal17}. 

For the first set of experiments, we compare GCN, GraphSage, and MLP against Mixhop~\cite{abu_etal19}. A Mixhop layer uses a higher-order aggregation method to aggregate features from beyond a node's immediate neighborhood, a mechanism that has been touted as particularly effective for learning over heterophilic graphs~\cite{Zhu_etal20}. We use the same synthetic graphs provided by the Mixhop authors which have 5000 nodes and 10 classes and a 33\%/33\%/33\% train/validate/test split.

\begin{figure*}[h]
  \centering
  \begin{subfigure}{0.4\textwidth}
    \includegraphics[width = \textwidth]{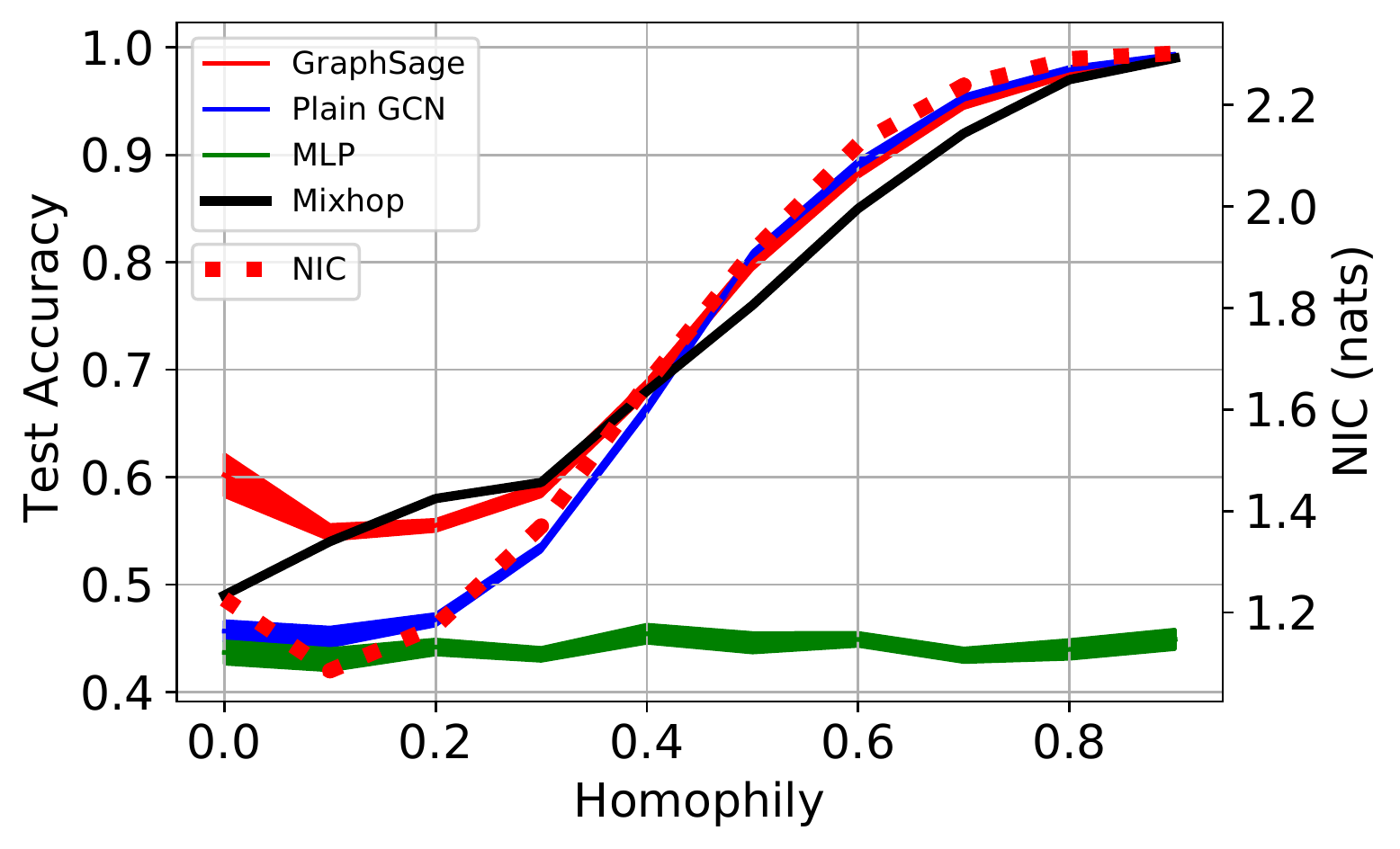} 
    \subcaption{}
    \label{fig:mixhop}
  \end{subfigure}
  \quad
  \begin{subfigure}{0.4\textwidth}
    \includegraphics[width = \textwidth]{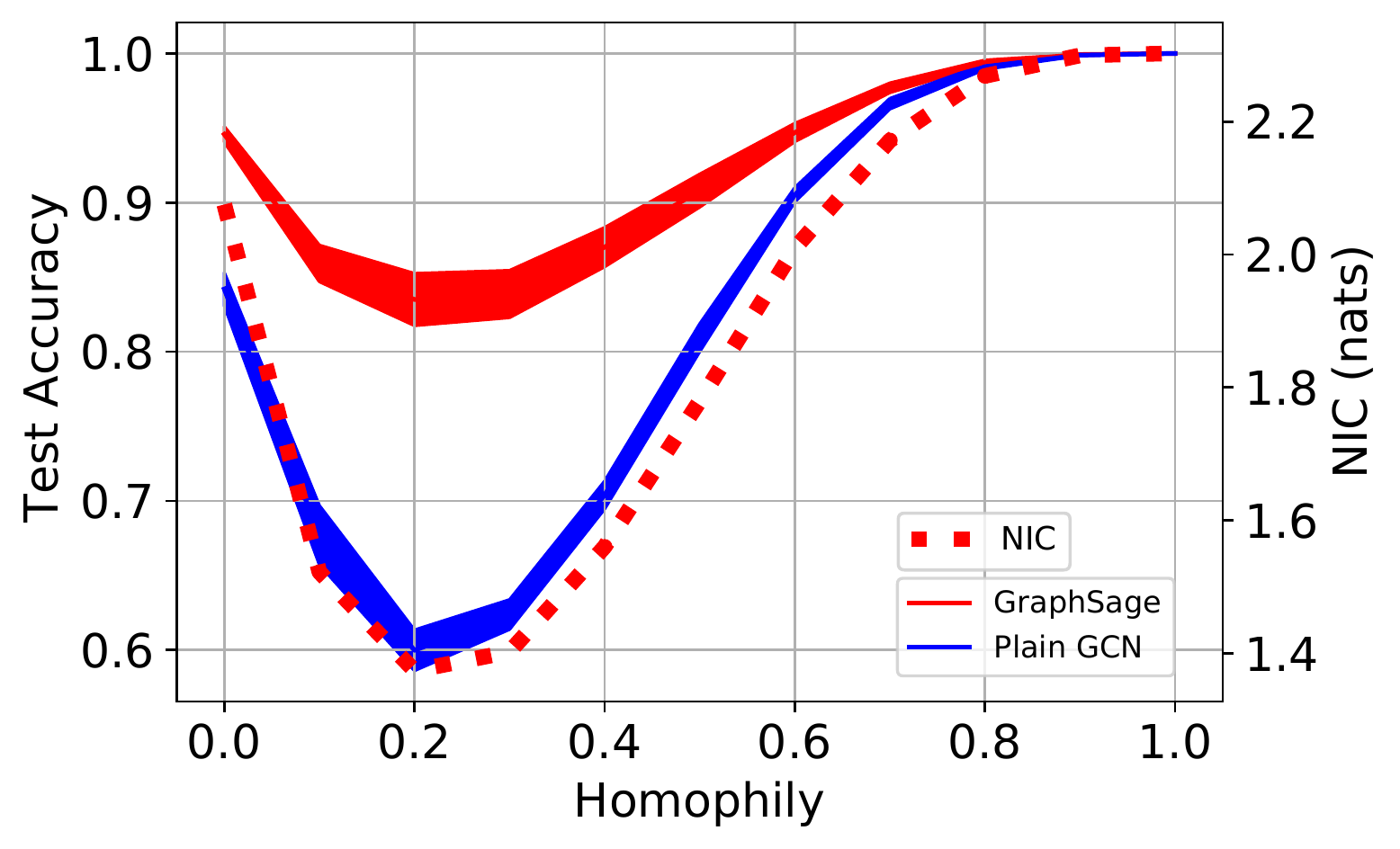} 
    \subcaption{}    
    \label{fig:syn}
  \end{subfigure}

  \caption{(\subref{fig:mixhop}) Test accuracy on the Mixhop synthetic graphs with variable homophily levels in the range $[0.0,0.9]$. The accuracy of Mixhop~\cite{abu_etal19} is taken from the original paper. The plot also shows the NIC metric as a function of the homophily of the underlying graph.
    (\subref{fig:syn}) Test accuracy and NIC on the syn-products synthetic graphs at homophily levels in the range $[0.0,1.0]$.}
  \label{fig:non_local_graph}
\end{figure*}

Figure~\ref{fig:mixhop} shows the accuracy of various models at different homophily levels. The mean and standard deviation (thickness of the line) at each homophily level is obtained from 10 trials. All models have two hidden layers. While the accuracy of Mixhop is better than GCNs at low homophily levels, its advantage disappears when we use GraphSage. Note that, unlike Mixhop, a GraphSage layer only aggregates features from a node's immediate neighborhood. We also plot the NIC metric for the graphs at different homophily levels. As shown in Fig.~\ref{fig:mixhop}, the NIC metric correlates better with the accuracy of GCNs than the homophily metric with a Pearson correlation coefficient of $0.997$ compared to $0.970$ for homophily.

We also test on the synthetic datasets from ref.~\cite{Zhu_etal20}. These graphs follow a similar generation process as ref.~\cite{Karimi_etal17} with a preferential attachment procedure that takes the node labels into account. The node features, however, are not randomly chosen, but copied from a real-world dataset such that features from nodes with the same label in the real-world dataset are mapped to nodes with the same label in the synthetic graph.
We follow the same generation procedure and generation parameters as ref.~\cite{Zhu_etal20} and use ogbn-products~\cite{Hu_etal20} as the source for node features.
As in ref.~\cite{Zhu_etal20}, we use a 25\%/25\%/50\% train/validate/test split. We run 10 trials at each homophily level.

Table~\ref{tab:syn_results} shows the results on the syn-products datasets together with the results of the $H_2$GCN models from ref.~\cite{Zhu_etal20}. Our GraphSage model significantly outperforms $H_2$GCN at low homophily levels and the two methods achieve similar accuracy at higher homophily levels. $H_2$GCN has multiple features that make it more powerful than GraphSage: it uses multi-hop aggregation within each layer to separately aggregate features from 1-hop and 2-hop neighborhoods, and uses skip connections to feed all intermediate representations to the top classifier. Table~\ref{tab:syn_results}, however, shows a similar situation to that in Table~\ref{tab:main_results} which is that these extra mechanisms are superfluous for the considered datasets.

Figure~\ref{fig:syn} plots the NIC metric for the syn-products graphs at the various homophily levels, together with the accuracy of GCN and GraphSage. Again, we see that NIC correlates better with GCN accuracy with a Pearson correlation coefficient of $0.987$ compared to $0.800$ for the homophily metric.

\setlength{\tabcolsep}{4pt}
\begin{table*}[h]
\caption{Mean percentage accuracy and standard deviation on the syn-products synthetic datasets}
\label{tab:syn_results}
\begin{center}
  \small
\begin{tabular}{l|ccccccc}
\hline
    $\mathcal{H}_{edge}$ & 0.0 & 0.1 & 0.2 & 0.3 & 0.4 & 0.5   \\
    \hline
    GCN & $0.85\pm 0.01$ & $0.68 \pm 0.02$ & $0.60 \pm 0.01$ & $0.62 \pm 0.01$ & $0.70 \pm 0.01$ & $0.81 \pm 0.01$  \\
    GraphSage & ${\bf 0.95 \pm 0.005}$ &  ${\bf 0.86 \pm 0.01}$  & ${\bf 0.83 \pm 0.02}$ & ${\bf 0.84 \pm 0.02}$  & ${\bf 0.87 \pm 0.01}$ & ${\bf 0.91 \pm 0.01}$  \\
    MLP & $0.67 \pm 0.02$ & $0.67 \pm 0.03$ & $0.67 \pm 0.03$ & $0.67 \pm 0.02$ & $0.67 \pm 0.03$ & $0.67 \pm 0.03$  \\
    $H_2$GCN-1 & $0.83 \pm 0.002$ & $0.78 \pm 0.02$ & $0.79 \pm 0.002$ & $0.81 \pm 0.002$ & $0.84\pm 0.01$ & $0.88 \pm 0.003$  \\
    $H_2$GCN-2 & $0.84 \pm 0.004$ & $0.80 \pm 0.008$ & $0.81 \pm 0.004$ & $0.83 \pm 0.005$ & ${\bf 0.87 \pm 0.007}$ & ${\bf 0.91 \pm 0.004}$ \\    
    \hline
    $\mathcal{H}_{edge}$ & 0.6 & 0.7 & 0.8 & 0.9 & 1.0  \\
    \hline
    \hline
    GCN & $0.91 \pm 0.004$ & $0.97 \pm 0.004$ & $0.99 \pm 0.002$ & ${\bf 1.00 \pm 3.0e-4}$ & ${\bf 1.00 \pm 6.0e-5}$  & \\
    GraphSage & ${\bf 0.95 \pm 0.006}$ & ${\bf 0.98 \pm 0.003}$ & ${\bf 0.99 \pm 0.001}$ & ${\bf 1.00 \pm 3.0e-4}$ & ${\bf 1.00 \pm 6.6e-5}$ & \\
    MLP & $0.67 \pm 0.03$ & $0.67 \pm 0.03$ & $0.67 \pm 0.03$ & $0.67 \pm 0.02$ & $0.67 \pm 0.02$ & \\
    $H_2$GCN-1 & $0.92 \pm 6.0e-4$ & $0.96 \pm 0.002$ & $0.98 \pm 0.02$ & ${\bf 1.00 \pm 0.001}$ & ${\bf 1.00 \pm 1.0e-4}$ &  \\
    $H_2$GCN-2 & ${\bf 0.95 \pm 0.003}$ & ${\bf 0.98 \pm 0.002}$ & ${\bf 0.99 \pm 5.0e-4}$ & ${\bf 1.00 \pm 8.0e-4}$ & ${\bf 1.00 \pm 1.0e-4}$ &  \\

\end{tabular}
\end{center}
\end{table*}

\subsection{Homophily and Accuracy in Node Classification Tasks}
\label{sec:homophily_analysis}
In this section, we investigate in more detail the relation between homophily and the accuracy of common graph neural network architectures such as GCNs and GraphSage. Common wisdom dictates that these architectures based on local aggregation would perform poorly on heterophilic datasets~\cite{Zhu_etal20,Pei_etal20,Mostafa_Nassar20,Liu_etal20}. We study this claim by using simple synthetic datasets with adjustable homophily levels. We generate the graphs in these datasets using the label-dependent preferential attachment method from ref.~\cite{Karimi_etal17} which is outlined in section~\ref{sec:synth-data}. In the generated graphs, we set the node features to be normally distributed vectors conditioned on node labels. Given a label $t_i=k$ for node $i$, the feature vector ${\bf h}^{0}_i$ has the distribution:
\begin{equation}
    P({\mathbf h}^{0}_i | t_i=k) = Gaussian({\mathbf h}^{0}_i; \mathbf{\mu}_k, \mathbf{\Sigma}_k) \quad \forall i \in \mathcal{V}
\end{equation}
where $Gaussian$ is the Gaussian probability density function. $\mathbf{\mu}_k$ and $\mathbf{\Sigma}_k$ are the label-dependent mean and covariance. This simple formulation ensures the node features carry some information about the node labels which is typically the case for real-world datasets. We generate graphs of $10,000$ nodes with 2 or 4 class labels and split them into a 20\%/10\%/80\% training/validation/testing split. We test the accuracy of GCNs, GraphSage, and MLPs. The latter ignores the graph structure. We also plot the single node maximum a-posteriori (MAP) detection bound (based only on the node's own feature). The MAP detection rule predicts label $\hat{t}_i$ for node $i$ according to: 
\begin{flalign*}
  \hat{t}_i =  \underset{k\in\mathcal{C}}{\arg\max}~[p(t_i = k|{\mathbf h}^{0}_i)]  =   \underset{k\in\mathcal{C}}{\arg\max}~[p({\mathbf h}^{0}_i|t_i = k)p(t_i=k)] 
\end{flalign*}
where $\mathcal{C}$ is the set of node labels. This rule reduces to a simple threshold operation for the scalar case and the MAP accuracy is given by 
\begin{math}
P_{MAP} = \sum_k p(t_i=k)Pr[\hat{t_i}=k|t_i=k].
\end{math}
The bound provides the best accuracy achievable by ignoring the graph structure.

\begin{figure*}[h]
  \centering
  \begin{subfigure}{0.4\textwidth}
    \includegraphics[width = \textwidth]{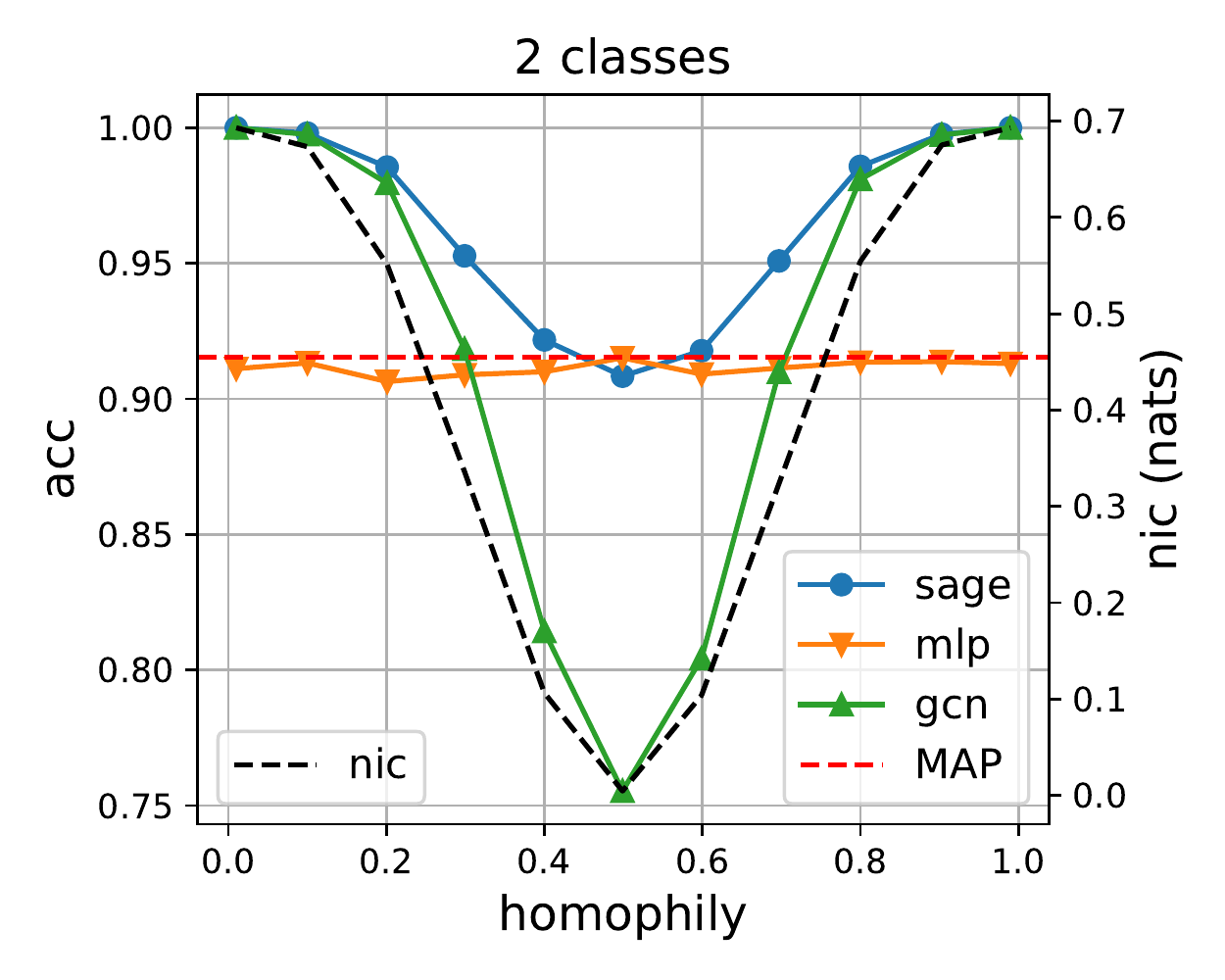} 
    \subcaption{}
    \label{fig:2cls-hom}
  \end{subfigure}
  \quad
  \begin{subfigure}{0.4\textwidth}
    \includegraphics[width = \textwidth]{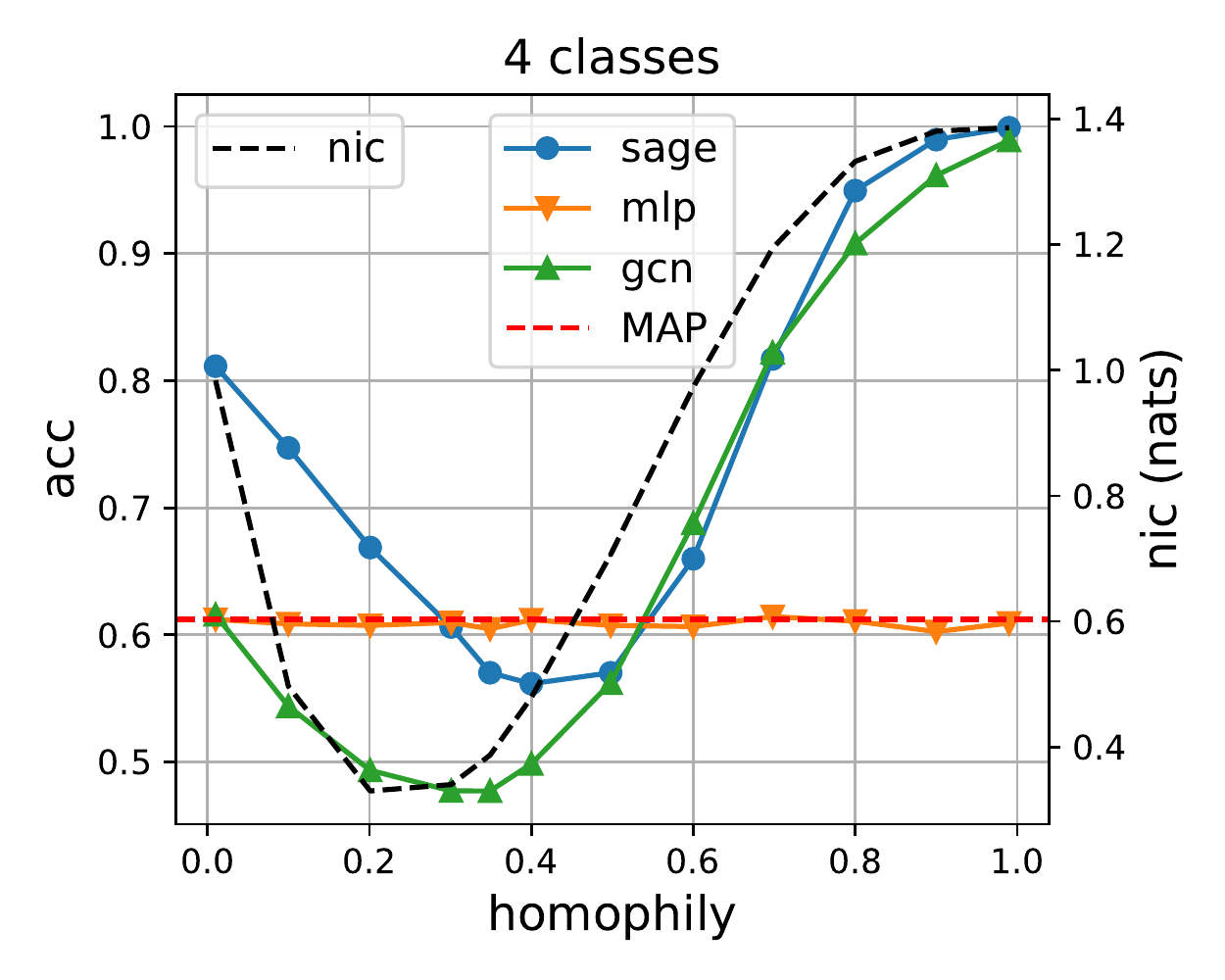} 
    \subcaption{}    
    \label{fig:4cls-hom}
  \end{subfigure}
  \\
  \begin{subfigure}{0.4\textwidth}
    \includegraphics[width = \textwidth]{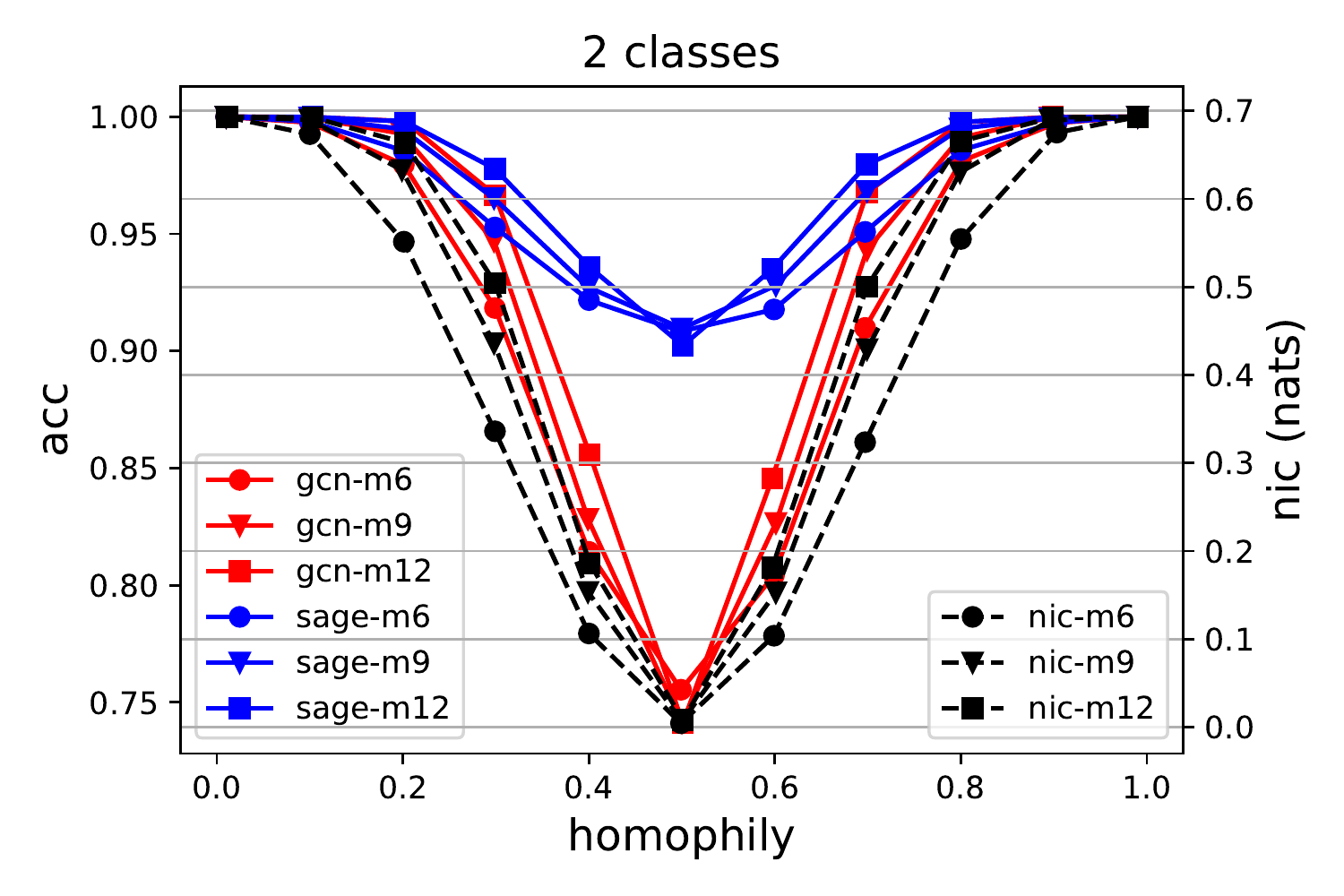} 
    \subcaption{}
    \label{fig:m-2cls}
  \end{subfigure}
  \quad
  \begin{subfigure}{0.4\textwidth}
    \includegraphics[width = \textwidth]{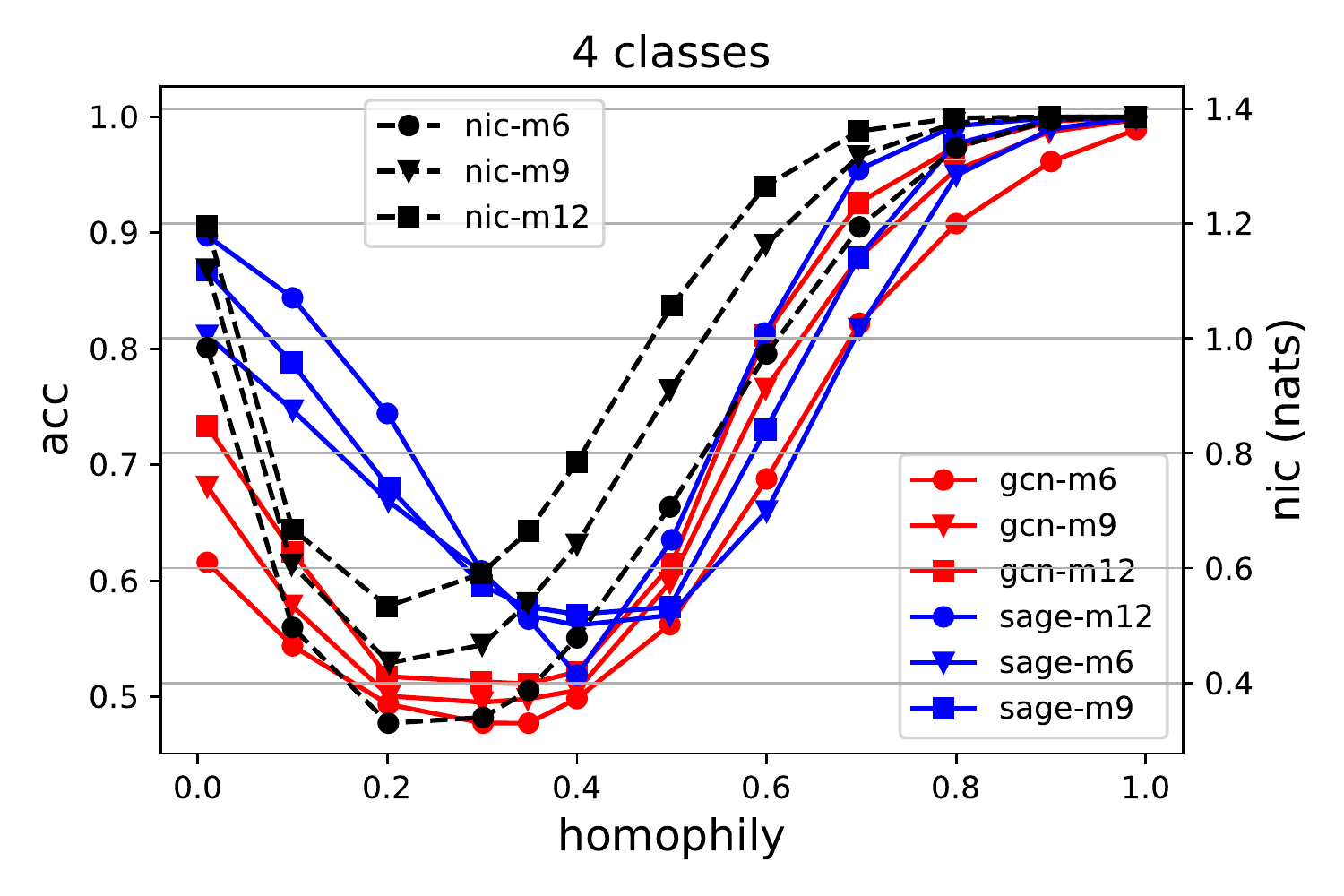} 
    \subcaption{}    
    \label{fig:m-4cls}
  \end{subfigure}
  
  \caption{(\subref{fig:2cls-hom})(\subref{fig:4cls-hom}) Test accuracy vs. homophily on datasets with 2 and 4 classes, respectively. 
    Pearson correlation coefficient between NIC and GCN accuracy is 0.948 in both \subref{fig:2cls-hom} and \subref{fig:4cls-hom} and is 0.0 and 0.790 between homophily and accuracy in \subref{fig:2cls-hom} and \subref{fig:4cls-hom},respectively. (\subref{fig:m-2cls})(\subref{fig:m-4cls}) Similar trends as in \subref{fig:2cls-hom} and \subref{fig:4cls-hom}. Note how the NIC metric depends on the connection density in the graph.
  }
  \label{fig:gnn_homophily}
\end{figure*}

Figure~\ref{fig:2cls-hom} shows the results for graphs with 2 node classes. As expected, the MAP bound is independent of the homophily value since it ignores the graph structure. Similarly, the MLP is independent of homophily and tracks the MAP bound. Interestingly, GCN shows a large accuracy degradation for a large range of homophily values around $0.5$ while attaining higher accuracy than MLP for very high and very low homophily values. Since GCN weighs both the node's own feature vector and neighboring feature vectors equally (see Eq.\ref{eq:gcn}), it is unable to learn to ignore those neighbors when they are uninformative about the node label. In the homophily range around 0.5, the neighbors appear as if they were randomly sampled and their aggregation message acts as a noise source at the node. This trend is later reversed at low homophily values as the node's label is almost guaranteed to be the complement of its neighbors (we are using binary labels) . It is thus straightforward to see that the accuracy of GCN at very high (1.0) and very low (0.0) homophily values should be identical as the label of a node can be inferred with high confidence from the label of only one of its neighbors.  

On the other hand, since GraphSage weighs the node's own feature vector separately from its neighbors' aggregation message (see Eq.~\ref{eq:gsage}), it is able to adapt to the neighborhood information content by acting like an MLP (with slight degradation in accuracy) in the low neighborhood information regime (around homophily value 0.5), while still being able to utilize the information in the neighborhood at low and high homophily values, when this information correlates more strongly with the node's own label. 

A similar trend holds for the 4 classes case as shown in Fig.~\ref{fig:4cls-hom}. Here, GCN reaches its lowest accuracy and underperforms MLP for the homophily range between 0.2 and 0.4. Again GraphSage  does a better job at balancing the contributions from a node's neighborhood and the contributions from the node's own feature vector depending on the information content in the neighborhood.

One shortcoming of the homophily metric is that it does not depend on the number of node classes. The number of possible node classes strongly affects the difficulty of the node classification task as shown in Figs.~\ref{fig:2cls-hom} and~\ref{fig:4cls-hom}. Our NIC metric naturally captures this effect; the NIC metric has its lowest value at the point where the neighborhood label configuration is maximally non-informative (has maximum entropy) which in these particular synthetic example corresponds to the point where \mbox{$homophily =  \frac{1}{\text{number of labels}}$} as this is the point where each neighboring node is equally likely to belong to any class. 

We perform another set of experiments where we vary one of the graph generation parameters to obtain graphs with different degree distributions. The generation parameter we vary, $m$, controls the number of connections created by each new node that is added to the graph during the sequential generation process. Larger $m$ values lead to more densely connected graphs. Figures~\ref{fig:m-2cls} and ~\ref{fig:m-4cls} show the accuracy of GCN and GraphSage for different homophily and $m$ values. GCN and GraphSage achieve better accuracy for larger $m$ values. The NIC metric is able to capture this effect and yields higher values for larger $m$ values. The homophily metric is insensitive to the graphs' degree distribution.

\section{Conclusions}
The field of GNNs has often grappled with non-standardized evaluation approaches that make it hard to evaluate the strengths of new methods. For the popular citation datasets (Cora, Citeseer, and Pubmed) for example, careful hyperparameter tuning and early stopping indicate that more elaborate GNN aggregation methods do not offer significant performance advantages~\cite{Shchur_etal18}. Our results indicate a similar situation exists in the area of heterophilic datasets. We ran experiments on a wide range of real-world and synthetic heterophilic datasets, and showed that simple models like MLPs, GCNs, and GraphSage perform on par with various recent GNN methods that were specifically designed to boost performance on heterophilic graphs. Part of the issue lies with the heterophilic datasets in common use as many of them are quite small (a few hundred nodes) which leads to widely different accuracy figures depending on the particular random split used. Another issue is the lack of standardized (non-random) training/validation/testing splits, and even a lack of agreement on the percentage of nodes to use for training, validation, and testing. Recently, there has been a shift away from using the Cora, Citeseer, and Pubmed graph datasets in favor of larger and more robust datasets~\cite{Hu_etal20}. Our results indicate a similar shift away from the commonly used heterophilic datasets might be necessary in order to find datasets and problem domains where the advantages of multi-hop and long-range aggregation methods are more apparent.

We show that label-dependent graph homophily is a poor metric for predicting the accuracy of GNNs in node classification tasks. While GNNs in general perform worse on graphs with low homophily, this trend does not hold in many cases. We show that in several cases, the accuracy of GNNs on graphs with lower homophily is actually better than on graphs with higher homophily. We proposed a more comprehensive metric that considers the entire distribution of labels in a node's 1-hop neighborhood.
By virtue of being an information-theoretic metric, our NIC metric is quite general and does not make any assumptions about the aggregation method used to collect information from the local neighborhood.

NIC takes into account several factors such as the distribution of labels, the connection probabilities between all label pairs, and the label-dependent degree distribution. We showed on several datasets that by taking these factors into account, NIC correlates better with GNN performance than homophily. NIC is also easy to evaluate using quantities that can be quickly and empirically estimated from the graph. NIC could thus be used to make more informed judgments about the suitability of GNNs with local aggregation to various node classification problems on graphs, and NIC could be used to construct or to find graphs where GNNs with local aggregation are expected to perform poorly. Such graphs could serve as more relevant benchmarks for long-range aggregation methods than the currently used heterophilic datasets.

\appendix
\section{Hyper-parameter selection}
\label{app:hyper}
All the networks we use (plain GCNs, GraphSage, and MLPs) have two hidden layers with the same dimensions. We use dropout between all layers. For all datasets we do a hyper-parameter sweep and use the hyper-parameter point with the best validation accuracy. For simplicity, the hyper-parameters are the same for all models (plain GCN, GraphSage, and MLP) running on a particular dataset. We choose the hyper-parameters that give the best validation accuracy on the best-performing model. We always train for 2000 iterations per trial and report test accuracy at the iteration with best validation accuracy for the trial. We sweep over the following hyper-parameters:
\begin{enumerate}
\item {\bf learning rate ($lr$)}: sweep over $\{0.005,0.05,0.1\}$
\item {\bf Dropout probability ($p_{drop}$)}: sweep from 0.0 to 0.7 in steps of 0.1
\item {\bf hidden layer size ($N_h$)}: sweep over $\{256,512,1024\}$
\item {\bf L2 weight decay ($WD$)} : sweep over $\{0.0,5.0e-5,5.0e-4\}$
\end{enumerate}
Table~\ref{tab:hyperparams} lists the hyper-parameter used for each dataset.

\begin{table*}[h]
\caption{Hyper-parameter choices for all datasets}
\label{tab:hyperparams}
\begin{center}
\begin{tabular}{l|cccccccc}
\hline
  & Chameleon & Squirrel  & Cornell & Texas & Wisconsin & Actor & Mixhop & syn-products \\
    \hline
    \hline
    $lr$ & 0.005  & 0.005 & 0.05 & 0.05 &  0.05 & 0.005 & 0.05 & 0.05 \\
    $p_{drop}$ & 0.6 & 0.6  & 0.4 & 0.4 & 0.4 & 0.5 & 0.1 & 0.5\\
    $N_h$ & 512 & 512  & 256 & 256 & 256 & 256 & 1024 & 512\\
    $WD$ & 0.0 & 0.0  & 5.0e-4 & 5.0e-4 & 5.0e-4 & 0.0 & 0.0 & 0.0\\
\end{tabular}
\end{center}
\end{table*}

\section{Proof of theorem 1}
\label{app:proof}

For notational simplicity, we denote the event that an edge exists from node $w$ to node $v$ as $e_{vw}$ and we denote the label of node $v$ as $l(v) \equiv l_v$.

We begin by deriving the probability distribution over a node's label given that it connects to another node with label $s$. 
\begin{flalign*}
  P(l_w = r | l_v =s , e_{vw})  = & \frac{P(e_{vw} | l_v = s,l_w = r) P(l_v = s,l_w = r)}{P(l_v=s,e_{vw})} \\
  = & \frac{c_{sr} P(l_v = s)P(l_w = r)}{P(e_{vw} | l_v = s)P(l_v=s)} \\
  = & \frac{c_{sr}p_r}{\sum\limits_k P(e_{vw},l_w = k | l_v=s)} \\
  = & \frac{c_{sr}p_r}{\sum\limits_k P(e_{vw} | l_v=s,l_w = k)P(l_w = k)} \\
  = & \frac{c_{sr}p_r}{\sum\limits_k c_{sk}p_k} \\
  = & z_{sr} \\      
\end{flalign*}
$l(\mathcal{N}(v)) = \{l_w : w \in \mathcal{N}(v)\}$ is the label configuration of the neighborhood of node $v$. Since the configurations are permutation-invariant, we are more interested in the number of occurrences of different labels in the neighborhood. Define $\mathcal{N}_{vs} = |\{l_w : w \in \mathcal{N}(v) \wedge l_w = s|\}$ as the number of nodes with label $s$ in the neighborhood of node $v$. Given the assumption that the number of nodes in the graph ($N$) is much larger than the maximum node degree ($D$),  the probability of observing a particular label configuration in the neighborhood of a node with label $s$ where the neighborhood has size $d$ is given by the multinomial(MN) distribution:
\begin{flalign}
  \label{eq:cond}
  P(l(\mathcal{N}\left(v)\right) | l_v = s) = & q_s(d)MN(\mathcal{N}_{v1},\ldots,\mathcal{N}_{vC};d;z_{s1},\ldots,z_{sC}) \notag \\
  = & q_s(d)\frac{d!}{\prod\limits_{k=1}^C\mathcal{N}_{vk}!} \prod\limits_{k=1}^C z_{sk}^{\mathcal{N}_{vk}}
\end{flalign}
where $q_s$ is the probability distribution over the degree of nodes with label $s$. Note that we need $N \gg D$ to ensure that we do not have to take into account the possible depletion of connection sources for large neighborhood sizes. The marginal distribution over neighborhood label configurations is:
\begin{equation}
\label{eq:mixture}
  P(l(\mathcal{N}\left(v)\right)) = \sum\limits_{s=1}^C p_s q_s(d) \frac{d!}{\prod\limits_{k=1}^C\mathcal{N}_{vk}!} \prod\limits_{k=1}^C z_{sk}^{\mathcal{N}_{vk}},
\end{equation}
where we simplify notation by always using $d$ in place of $|\mathcal{N}(v)|$. We can write the mutual information between the label of a node and the label configuration of its neighborhood as:
\begin{equation}
  \label{eq:MI}
  MI(l(\mathcal{N}\left(\mathcal{V})\right),l_{\mathcal{V}}) = H(l(\mathcal{N}\left(\mathcal{V})\right)) - H(l(\mathcal{N}\left(\mathcal{V})\right) | l_{\mathcal{V}} ),
\end{equation}
The second term on the right is the entropy of the conditional distribution defined in Eq.~\ref{eq:cond} which can easily be obtained from the standard expression for the entropy of the multinomial. The first term is the entropy of the marginal distribution defined in Eq.~\ref{eq:mixture}. The marginal has the form of a mixture distribution. The entropy of mixture distributions often has no closed form, even for simple mixture components such as Gaussian components. We thus resort to using a lower bound on the entropy of the mixture distribution defined in Eq.~\ref{eq:mixture}. We use the result from ref.~\cite{Kolchinsky_Tracey17} which we reproduce here:

\begin{lemma} 
  \label{lemma:ref}
  Assume we have a distribution $p_L(l)$ over $C$ possible outcomes, where $p_s = p_L(l=s)$. Consider the mixture distribution:
  $$z_{X}(x) = \sum\limits_{s=1}^C p_s z_s(x),$$
  where $z_s$ is the probability density of component $s$, we have the following lower bound on the entropy $H(X)$ of $z_X(x)$:
  \begin{equation*}
    H(X) \geq H(X|L) - \sum\limits_{s=1}^Cp_s ln\left(\sum\limits_{r=1}^C p_r BD(z_s || z_r)\right),
  \end{equation*}
  where
  \begin{equation*}
    BD(z_1 || z_2) = \sum\limits_x \sqrt{z_1(x)z_2(x)}
  \end{equation*}
\begin{proof}
See ref.~\cite{Kolchinsky_Tracey17}
\end{proof}
\end{lemma}
$BD$ is the Bhattacharyya or expected likelihood kernel~\cite{Jebara_Kondor03}. The mixture components in our case are multinomials. These are the conditional distributions from Eq.~\ref{eq:cond}. The Bhattacharyya distance between two mixture components in our case is:
\begin{flalign}
  BD & \left(P(l(\mathcal{N}\left(v)\right) | l_v = s) || P(l(\mathcal{N}\left(v)\right) | l_v = r)\right) \notag \\
  = & \sum\limits_{d=1}^D\sum\limits_{\mathcal{N}_{v1} + \ldots + \mathcal{N}_{vC} = d}\frac{\sqrt{q_s(d)q_r(d)}d!}{\prod\limits_{k=1}^C\mathcal{N}_{vk}!}\prod\limits_{k=1}^C (z_{sk}z_{rk})^{\mathcal{N}_{vk}/2} \notag \\
  = & \sum\limits_{d=1}^D\sqrt{q_s(d)q_r(d)}\left(\sum\limits_{k=1}^C \sqrt{z_{sk}z_{rk}}\right)^d \label{eq:bat},
\end{flalign}
where the last step follows from the multinomial theorem. Using lemma ~\ref{lemma:ref} and the expression for the Bhattacharyya distance from Eq.~\ref{eq:bat}, the entropy of the marginal distribution in Eq.~\ref{eq:mixture} is bounded by:
\begin{flalign*}
  H(l(\mathcal{N}\left(\mathcal{V})\right)) \geq &  H(l(\mathcal{N}\left(\mathcal{V})\right) | l_{\mathcal{V}} ) \\
  - & \sum\limits_{s=1}^Cp_s ln\left(\sum\limits_{r=1}^C p_r\left( \sum\limits_{d=1}^D\sqrt{q_s(d)q_r(d)}\left(\sum\limits_{k=1}^C \sqrt{z_{sk}z_{rk}}\right)^d\right)\right) \label{eq:bat}
\end{flalign*}
By substituting this bound on $H(l(\mathcal{N}\left(\mathcal{V})\right))$ in the definition of MI in Eq.~\ref{eq:MI}, we obtain theorem~\ref{th:main}.


\begin{thebibliography}{10}

\bibitem{Kipf_Welling16}
T.N. Kipf and Max Welling.
\newblock Semi-supervised classification with graph convolutional networks.
\newblock {\em International Conference on Learning Representations}, 2017.

\bibitem{Shi_etal20}
Yunsheng Shi, Zhengjie Huang, Shikun Feng, and Yu~Sun.
\newblock Masked label prediction: Unified massage passing model for
  semi-supervised classification.
\newblock {\em arXiv preprint arXiv:2009.03509}, 2020.

\bibitem{Li_etal20}
Guohao Li, Chenxin Xiong, Ali Thabet, and Bernard Ghanem.
\newblock Deepergcn: All you need to train deeper gcns.
\newblock {\em arXiv preprint arXiv:2006.07739}, 2020.

\bibitem{Corso_etal20}
Gabriele Corso, Luca Cavalleri, Dominique Beaini, Pietro Li{\`o}, and Petar
  Veli{\v{c}}kovi{\'c}.
\newblock Principal neighbourhood aggregation for graph nets.
\newblock {\em arXiv preprint arXiv:2004.05718}, 2020.

\bibitem{Li_etal19}
Yujia Li, Chenjie Gu, Thomas Dullien, Oriol Vinyals, and Pushmeet Kohli.
\newblock Graph matching networks for learning the similarity of graph
  structured objects.
\newblock In {\em International Conference on Machine Learning}, pages
  3835--3845. PMLR, 2019.

\bibitem{Zhang_Chen18}
Muhan Zhang and Yixin Chen.
\newblock Link prediction based on graph neural networks.
\newblock {\em arXiv preprint arXiv:1802.09691}, 2018.

\bibitem{Newman_03}
M.E.J. Newman.
\newblock Mixing patterns in networks.
\newblock {\em Physical review E}, 67(2):026126, 2003.

\bibitem{Zhu_etal20}
Jiong Zhu, Yujun Yan, Lingxiao Zhao, Mark Heimann, Leman Akoglu, and Danai
  Koutra.
\newblock Beyond homophily in graph neural networks: Current limitations and
  effective designs.
\newblock {\em Advances in Neural Information Processing Systems}, 33, 2020.

\bibitem{Pei_etal20}
Hongbin Pei, Bingzhe Wei, Kevin Chen-Chuan Chang, Yu~Lei, and Bo~Yang.
\newblock Geom-gcn: Geometric graph convolutional networks.
\newblock In {\em International Conference on Learning Representations}, 2020.

\bibitem{abu_etal19}
Sami Abu-El-Haija, Bryan Perozzi, Amol Kapoor, Nazanin Alipourfard, Kristina
  Lerman, Hrayr Harutyunyan, Greg~Ver Steeg, and Aram Galstyan.
\newblock Mixhop: Higher-order graph convolutional architectures via sparsified
  neighborhood mixing.
\newblock In {\em international conference on machine learning}, pages 21--29.
  PMLR, 2019.

\bibitem{Mostafa_Nassar20}
Hesham Mostafa and Marcel Nassar.
\newblock Permutohedral-gcn: Graph convolutional networks with global
  attention.
\newblock {\em arXiv preprint arXiv:2003.00635}, 2020.

\bibitem{Liu_etal20}
Meng Liu, Zhengyang Wang, and Shuiwang Ji.
\newblock Non-local graph neural networks.
\newblock {\em arXiv preprint arXiv:2005.14612}, 2020.

\bibitem{Defferrard_etal16}
Micha{\"e}l Defferrard, Xavier Bresson, and Pierre Vandergheynst.
\newblock Convolutional neural networks on graphs with fast localized spectral
  filtering.
\newblock {\em arXiv preprint arXiv:1606.09375}, 2016.

\bibitem{Velivckovic_etal17}
Petar Veli{\v{c}}kovi{\'c}, Guillem Cucurull, Arantxa Casanova, Adriana Romero,
  Pietro Lio, and Yoshua Bengio.
\newblock Graph attention networks.
\newblock {\em arXiv preprint arXiv:1710.10903}, 2017.

\bibitem{Hamilton_etal17}
Will Hamilton, Zhitao Ying, and Jure Leskovec.
\newblock Inductive representation learning on large graphs.
\newblock In {\em Advances in neural information processing systems}, pages
  1024--1034, 2017.

\bibitem{Xu_etal18}
Keyulu Xu, Weihua Hu, Jure Leskovec, and Stefanie Jegelka.
\newblock How powerful are graph neural networks?
\newblock {\em arXiv preprint arXiv:1810.00826}, 2018.

\bibitem{Simonovsky_Komodakis17}
Martin Simonovsky and Nikos Komodakis.
\newblock Dynamic edge-conditioned filters in convolutional neural networks on
  graphs.
\newblock In {\em Proceedings of the IEEE conference on computer vision and
  pattern recognition}, pages 3693--3702, 2017.

\bibitem{Li_etal18b}
Qimai Li, Zhichao Han, and Xiao-Ming Wu.
\newblock Deeper insights into graph convolutional networks for semi-supervised
  learning.
\newblock In {\em Thirty-Second AAAI Conference on Artificial Intelligence},
  2018.

\bibitem{Chen_etal19}
Deli Chen, Yankai Lin, Wei Li, Peng Li, Jie Zhou, and Xu~Sun.
\newblock Measuring and relieving the over-smoothing problem for graph neural
  networks from the topological view.
\newblock {\em arXiv preprint arXiv:1909.03211}, 2019.

\bibitem{Klicpera_etal18}
Johannes Klicpera, Aleksandar Bojchevski, and Stephan G{\"u}nnemann.
\newblock Predict then propagate: Graph neural networks meet personalized
  pagerank.
\newblock {\em arXiv preprint arXiv:1810.05997}, 2018.

\bibitem{Joe89}
Harry Joe.
\newblock Estimation of entropy and other functionals of a multivariate
  density.
\newblock {\em Annals of the Institute of Statistical Mathematics},
  41(4):683--697, 1989.

\bibitem{Belghazi_etal18}
M.I. Belghazi, Aristide Baratin, Sai Rajeshwar, Sherjil Ozair, Yoshua Bengio,
  Aaron Courville, and Devon Hjelm.
\newblock Mutual information neural estimation.
\newblock In {\em International Conference on Machine Learning}, pages
  531--540. PMLR, 2018.

\bibitem{Tschannen_etal19}
Michael Tschannen, Josip Djolonga, P.K. Rubenstein, Sylvain Gelly, and Mario
  Lucic.
\newblock On mutual information maximization for representation learning.
\newblock {\em arXiv preprint arXiv:1907.13625}, 2019.

\bibitem{Kingma_Ba14}
D.P. Kingma and Jimmy Ba.
\newblock Adam: A method for stochastic optimization.
\newblock {\em arXiv preprint arXiv:1412.6980}, 2014.

\bibitem{Barabasi_Albert99}
Albert-L{\'a}szl{\'o} Barab{\'a}si and R{\'e}ka Albert.
\newblock Emergence of scaling in random networks.
\newblock {\em science}, 286(5439):509--512, 1999.

\bibitem{Karimi_etal17}
Fariba Karimi, Mathieu G{\'e}nois, Claudia Wagner, Philipp Singer, Strohmaier,
  and Markus.
\newblock Visibility of minorities in social networks.
\newblock {\em arXiv preprint arXiv:1702.00150}, 2017.

\bibitem{Hu_etal20}
Weihua Hu, Matthias Fey, Marinka Zitnik, Yuxiao Dong, Hongyu Ren, Bowen Liu,
  Michele Catasta, and Jure Leskovec.
\newblock Open graph benchmark: Datasets for machine learning on graphs.
\newblock {\em arXiv preprint arXiv:2005.00687}, 2020.

\bibitem{Shchur_etal18}
Oleksandr Shchur, Maximilian Mumme, Aleksandar Bojchevski, and Stephan
  G{\"u}nnemann.
\newblock Pitfalls of graph neural network evaluation.
\newblock {\em arXiv preprint arXiv:1811.05868}, 2018.

\bibitem{Kolchinsky_Tracey17}
Artemy Kolchinsky and B.D. Tracey.
\newblock Estimating mixture entropy with pairwise distances.
\newblock {\em Entropy}, 19(7):361, 2017.

\bibitem{Jebara_Kondor03}
Tony Jebara and Risi Kondor.
\newblock Bhattacharyya and expected likelihood kernels.
\newblock In {\em Learning theory and kernel machines}, pages 57--71. Springer,
  2003.

\end{thebibliography}
\end{document}